\documentclass[sigconf]{acmart}
\usepackage{cancel}
\usepackage[T1]{fontenc}
\usepackage[outline]{contour}
\usepackage{xcolor}

\AtBeginDocument{%
  \providecommand\BibTeX{{%
    \normalfont B\kern-0.5em{\scshape i\kern-0.25em b}\kern-0.8em\TeX}}}

\setcopyright{acmcopyright}
\copyrightyear{2018}
\acmYear{2018}
\acmDOI{10.1145/1122445.1122456}

\acmConference[Woodstock '18]{Woodstock '18: ACM Symposium on Neural
  Gaze Detection}{June 03--05, 2018}{Woodstock, NY}
\acmBooktitle{Woodstock '18: ACM Symposium on Neural Gaze Detection,
  June 03--05, 2018, Woodstock, NY}
\acmPrice{15.00}
\acmISBN{978-1-4503-9999-9/18/06}

\begin{document} 

\title{Heterogeneous Causal Learning for Effectiveness Optimization in User Marketing} 



\author{Will Y. Zou}
\email{will.zou@uber.com}
\affiliation{%
  \institution{Uber Inc.}
  \streetaddress{1455 Market Street}
  \city{San Francisco}
  \state{CA}
  \country{USA}
}

\author{Shuyang Du}
\email{shuyangdu@uber.com} 
\affiliation{%
  \institution{Uber Inc.}
  \streetaddress{1455 Market Street}
  \city{San Francisco}
  \state{CA}
  \country{USA}
}

\author{James Lee} 
\email{jameslee@uber.com} 
\affiliation{%
  \institution{Uber Inc.}
  \streetaddress{1455 Market Street}
  \city{San Francisco}
  \state{CA}
  \country{USA}
}

\author{Jan Pedersen} 
\email{jpedersen@uber.com}
\affiliation{%
  \institution{Uber Inc.}
  \streetaddress{1455 Market Street}
  \city{San Francisco}
  \state{CA}
  \country{USA}
}


\begin{abstract} 
User marketing is a key focus for consumer based internet companies. Learning algorithms are effective to optimize marketing campaigns which increase user engagement, and facilitates cross-marketing to related products. By attracting users with rewards, marketing methods are effective to boost user activity in the desired products. Rewards incur significant cost that can be off-set by increase in future revenue. Most methodologies rely on churn predictions to prevent losing users to make marketing decisions, which cannot capture up-lift across counterfactual outcomes with business metrics. Other predictive models are capable of estimating heterogeneous treatment effects, but fail to capture the balance of cost versus benefit. 

We propose a treatment effect optimization methodology for user marketing. This algorithm learns from past experiments and utilize novel optimization methods to optimize cost efficiency with respect to user selection. The method optimizes decisions using deep learning optimization models to treat and reward users, which is effective in producing cost-effective, impactful marketing campaigns. Our methodology demonstrates superior algorithmic flexibility with integration with deep learning methods and dealing with business constraints. The effectiveness of our model surpasses quasi-oracle estimation (R-learner) model and causal forests. We also established evaluation metrics that reflect the cost-efficiency and real-world business value. 

Our proposed constrained and direct optimization algorithms outperform by 24.6\% compared with best performing method in prior art and baseline methods. The methodology is useful in many product scenarios such as optimal treatment allocation and it has been deployed in production world-wide.


\end{abstract} 






\maketitle 

\section{Introduction}
\label{sec:intro} 
Improving user marketing efficacy have become an important focus for many internet companies.  Customer growth and engagement are critical in a fast-changing market, and cost of acquiring new users are rising. New product areas are especially pressured to acquire customers. In different industries, companies provide various ways for user marketing and cross-sell to new products, examples include  ride-sharing (Uber, Lyft), accommodation (Airbnb), and e-commerce (Amazon, Ebay).

As suggested in previous research [1] from Uber, providing a user with a reward without explicit apology after an unsatisfactory trip experience will have a positive treatment effect on future billings. This is consistent with the finding in [2] where researchers conducted a similar experiment on Via (a ride-sharing company in NYC). Marketing campaigns in internet companies offer similar rewards to encourage users to engage or use new products. The treatment has positive effects on desired business growth, also lead to a surplus in cost. To study the outcome of these rewards, the research perspective originates from treatment effect  estimation~\cite{rubin1974estimating} in a population or users. Previous research and common practice relies on non-causal churn prediction or heuristics based on frustrating experiences for reward decisions instead of directly optimizing for users' treatment effects under a cost constraint. In this paper, we apply the treatment effect estimation perspective on user marketing scenarios. 

The goal of our work is provide a business decision methodology to optimize for the effectiveness of treatments. This methodology has the combined effect of minimizing cost and creating uplift in user engagement. Compared to existing work, novel contributions of this paper are: 

\begin{itemize}

\item \textbf{Heterogeneous Treatment Effect based Business Decisions} - A common approach for user reward decisions relies on regular predictions, redemption or heuristics which are tied to specific scenario and require rich background context. In this paper we propose a general methodology that directly optimizes the heterogeneous treatment effect and could be applied to various business use cases with minimum change. This approach can be evaluated effectively and give guidance to decisions. 

\item \textbf{Cost versus Benefit for Aggregated Efficiency} - Most research studies focus on treatment effect of one single outcome. However, in real-world applications it’s necessary to consider treatment effect on the cost, i.e. the efficiency ratio of  $\delta$cost/$\delta$value when making the resource allocation decision. Common approach also only considers point estimates but our objective is to maximize effectiveness from aggregated treatment effect. Our proposed framework will solve these two challenges together.

\item \textbf{Deep Learning Integration and Joint Objective} - Previous methodology have focused on greedily estimating the treatment effect across multiple outcomes. Their algorithmic approach rely on statistical regression methods or linear models. We develop methodologies that incorporate various dimensions of outcomes in the learning objective, so a desired, holistic metric can be optimized through deep learning. This makes the algorithm flexible to integrate with deep learning algorithms. 

\item \textbf{Barrier Function for Constrained Optimization} - Constraints such as budgets, geography limitations,  affect user behavior in sophisticated ways. User state variations under barrier constraints form a novel problem space. We formulate a constrained ranking algorithm to learn combined effect of actions and constraints in production. This is a all-purpose model that can be used to model both market-wide efficiency, and treatment effects with limited resources.
\end{itemize} 

The structure of this paper is as follows: in Section~\ref{sec:related_work}, we will cover related work in optimization of treatment effect. In Section~\ref{sec:algorithms}, we make the problem statement and introduce effectiveness measures and our modeling approaches for treatment effect optimization. In Section~\ref{sec:empirical_results}, we will cover experimentation, results, comparisons across models and real-world performance from the product we launched. Finally we briefly cover future research steps. 
\vspace{-0.2cm}
\section{Background} 
\label{sec:related_work} 
Methods optimizing for user marketing, rewards and retention have been widely studied. Two recent studies by Halperin et al. [1] and Cohen et al. [2]  look into the effect of apology treatments when the user's trust is compromised. Andrews et al. [19] studied factors that affect coupon redemption. Hanna et al. [20] and Manzoor and Akoglu [21] investigated factors that influence redemption of time limited incentives. These studies focus on redemption or exploratory average treatment effect and do not explore the optimization of user selection. 

The above methods attempt to solve the business problem, and do not yet apply a causal learning approach. ~\cite{rubin1974estimating} first brought forward a framework for studying treatment effects. User instances are treated with an action, and when the outcome is observed it is used in model fitting. One significant area is application of statistical methods such as~\cite{kunzel2017meta} that decomposes the learning algorithm into composite models with \emph{meta-learners}. The study of \emph{meta-learners} have developed to a variety of models. Another area is application of decision trees and random forests~\cite{chen2016xgboost}, for instance, uplift tree\cite{rzepakowski2012decision}, causal tree and random forests~\cite{wager2017estimation}~\cite{athey2016recursive}, boosting~\cite{powers2017some} are powerful components to build causal inference models. Recently, another widely-adopted framework for learning heterogeneous treatment effect is the work of quasi-oracle estimation by~\cite{nie2017quasi}, which is proven to be effective when estimating the treatment effect in a single outcome. These methods consider both the \emph{Conditional Treatment Effect (CTE)} and the \emph{Average Treatment Effect (ATE)}. The \emph{CTE} is the treatment effect predicted by the model per sample conditional on its features while \emph{ATE} is the overall treatment effect. However, these algorithms are designed to estimate \emph{ATE} and \emph{CTE} for single outcome but could not deal with multiple outcomes and benefit-cost trade off. In this work we propose a set of algorithms which not only able to predict effect of treatment, but combine multiple outcomes into effectiveness measures that can be optimized jointly. 

\subsection{Estimation of Treatment Effect} 
We start with the estimation of treatment effects with the potential outcomes framework (Neyman, 1923~\cite{neyman23thesis}; Rubin, 1974~\cite{rubin1974estimating}) consistent with prior work~\cite{nie2017quasi}. In the user retention case, users are $n$ independent and identically distributed examples indexed by $i$, where $\mathbf{X}^{(i)}$ denotes per-sample features for user $i$ while $\mathbf{X}$ is the entire dataset, $Y_1^{(i)}$ is the observed outcome if treated, and $Y_0^{(i)}$ is observed outcome if not treated. $T^{(i)}$ is the treatment assignment and is binary for a particular treatment type, i.e. $T^{(i)} \in \{0, 1\}$. 

We assume the treatment assignment is unconfounded, i.e., the outcome pair is independent of treatment label given the user features, or treatment assignment is as good as random once we control for the features  (Rosenbaum and Rubin, 1983~\cite{Rosenbaum83propensity}): $\{Y_0, Y_1\} \perp T_i|\mathbf{X}_i$. This is the assumption we make on all causal models we explore in the paper. The treatment propensity, probability of a user receiving treatment as $e(\mathbf{x}^{(i)}) = P (T = 1  | \mathbf{X}^{(i)} = \mathbf{x}^{(i)})$. 

With experiments, outcomes are observed given the treatment assignments. With each user we would have only observed one outcome per treatment. This historical data can be used to fit a model. For treatment effect estimation, we seek to estimate the treatment effect function given that we observe user features $X$: 
\begin{align} 
\tau^*(\mathbf{x}) = E(Y_1 - Y_0 | \mathbf{X} = \mathbf{x})
\end{align} 
\vspace{-0.3cm}

\subsection{Quasi-oracle Estimation (R-learner)} 
Closely related to our work, we briefly review of the quasi-oracle estimation algorithm~\cite{nie2017quasi} for heterogeneous treatment effects, also known as \emph{`R-learner'}. The quasi-oracle estimation algorithm is a two-step algorithm for observational studies of treatment effects. The marginal effects and treatment propensities are first evaluated to form an objective function that isolates the causal component of the signal. Then the algorithm optimizes for the up-lift or causal component using regression. 

Concretely, the conditional mean of outcomes giving user features are $\mu^{*}_T(\mathbf{x}) = E(Y_T | \mathbf{X} = \mathbf{x})$, thus expected value of outcome from the model is $E(Y_T^{(i)} | \mathbf{X}^{(i)}) = E(Y_0^{(i)} | \mathbf{X}^{(i)}) + T^{(i)} E(Y_1^{(i)} - Y_0^{(i)} | \mathbf{X}^{(i)}) = \mu^*_0(\mathbf{X}^{(i)}) + T^{(i)}\tau^*(\mathbf{X}^{(i)})$. The expected value of the error $\epsilon$ across data and expected value of Y is zero given unconfoundedness assumption: \vspace{-0.2cm}
\begin{align} 
E(\epsilon(T^{(i)}) | \mathbf{X}^{(i)}, T^{(i)}) = 0
\end{align} 
\vspace{-0.1cm}
Replacing $E(Y_T^{(i)})$ in the error, and substitute the conditional mean outcome: $\epsilon(T^{(i)}) = Y_T^{(i)} - E(Y_T^{(i)}) = Y_T^{(i)} - (\mu^*_0(\mathbf{X}^{(i)}) + T^{(i)}\tau^*(\mathbf{X}^{(i)}))$; $m^*(\mathbf{x}^{(i)}) = E(Y^{(i)} | \mathbf{X}^{(i)} = \mathbf{x}^{(i)})$, we arrive at the decomposition: 
\vspace{-0.2cm}
\begin{align}
\label{eq:quasi_oracle_balance}
Y^{(i)} - m^*(\mathbf{X}^{(i)}) = (T^{(i)} - e^*(\mathbf{X}^{(i)})\tau^*(\mathbf{X}^{(i)}) + \epsilon
\end{align}
\vspace{-0.1cm} 
An equation that balances difference between outcome with a `mean' model with the conditional average treatment effect function. In a simple formulation of the quasi-oracle estimation algorithm a regression is used to fit the $m^*$ and $e^*$ models as the first step. The prediction result of the regression is then used to determine the regression target of $\tau^*$ model, which is then fitted also as a regression. After the learning, $\tau^*$ function can be used to estimate the treatment effect given user with feature $\mathbf{X}^{(i)}$. 

\section{Algorithms} 
\label{sec:algorithms} 
The quasi-oracle estimation algorithm is efficient for estimating conditional treatment effects, however, sometimes different outcomes incurred by treatment cannot be converted to the same unit, for example if we want to boost trip growth by increasing the dollar spend on rewards, trip number and dollar spend cannot be converted to a single value. So the eventual goal is to maximize gains and with a cost constraint. In this paper, we propose causal inference paradigm to maximize cost effectiveness of heterogeneous treatments. 

Concretely, we make the problem statement. Instead of estimating the treatment effect function $\tau^*(\mathbf{x}) = E(Y_1 - Y_0 | \mathbf{X} = \mathbf{x})$, we propose to solve the problem illustrated below to maximize the gain outcome given a cost constraint. 
\vspace{-0.1cm}
\begin{align} 
\label{eq:pstatement_constrained} 
\begin{split}
  \text{maximize} \quad &\sum_{i=1}^n\tau^{*r}(\mathbf{x}^{(i)})z_i \\ 
  \text{subject to} \quad &\sum_{i=1}^n\tau^{*c}(\mathbf{x}^{(i)})z_i \leq B \\
	&z_i \in\{0, 1\} 
\end{split}
\end{align} 
The variables $z_i$ represent whether we offer a reward to the user during a campaign and $B$ is the cost constraint. We represent retention treatment effects as $\tau^{*r}(\mathbf{x}^{(i)}) = E(Y_1^r - Y_0^r | \mathbf{X}^{(i)} = \mathbf{x}^{(i)})$ and cost as $\tau^{*c}(\mathbf{x}^{(i)}) = E(Y_1^c - Y_0^c | \mathbf{X}^{(i)} = \mathbf{x}^{(i)})$. It is important to note these treatment effect values are part of the optimization objective and are implicitly modeled as intermediate quantities. They are not strictly regression functions, and we holistically solve the stated problem. 
\vspace{-0.2cm}
\subsection{Duality R-learner} 
We describe the duality method with Lagrangian multipliers to solve the constrained optimization problem for maximizing gain (minimizing negative gain) subject to a budget ($B>0$) constraint, and relaxing the previous $z_i\in\{0, 1\}$ variables to continuous: 
\vspace{-0.1cm}
\begin{align} 
\label{eq:constrained_problem} 
\begin{split}
  \text{minimize} \quad -&\sum_{i=1}^n\tau^{*r}(\mathbf{x}^{(i)})z_i \\ 
  \text{subject to} \quad &\sum_{i=1}^n\tau^{*c}(\mathbf{x}^{(i)})z_i \leq B \\
	&\ 0\leq z_i \leq 1 
\end{split}
\end{align} 
\vspace{-0.1cm}
First, we assume the CATE functions are fixed, so we solve Problem~\ref{eq:pstatement_constrained} assuming $\tau^{*r}(\mathbf{x}^{(i)})$ and $\tau^{*c}(\mathbf{x}^{(i)})$ are given. Applying one Lagrangian multiplier, the Lagrangian for Problem~\ref{eq:pstatement_constrained}: 
\vspace{-0.2cm}
\begin{equation} 
  \label{eq:lagrangian} 
  L(\mathbf{z}, \lambda)=-\sum_{i=1}^n\tau^{*r}(\mathbf{x}^{(i)})z_i + \lambda (\sum_{i=1}^n\tau^{*c}(\mathbf{x}^{(i)})z_i - B) \\ 
\end{equation} 
\vspace{-0.1cm}
The optimization in Problem~\ref{eq:pstatement_constrained} can then be rewritten in its Dual form to maximize the Lagrangian dual function $g = \inf_{\mathbf{z}\in\emph{D}}L(\mathbf{z}, \lambda)$: 
\begin{align} 
\label{eq:duality_problem} 
\begin{split}
  \max\limits_{\lambda} \inf_{\mathbf{z} \in \emph{D}} L(\mathbf{z}, \lambda) \quad
  \text{subject to} \ 0\leq z_i \leq 1, \lambda \geq 0 \\
\end{split}
\end{align} 
We need to address the caveats for solving the problem with duality, and determine whether the dual problem has the same minimum with original problem.
\begin{itemize} 
\item If $p(\mathbf{z}, \lambda) = -\sum_{i=1}^n\tau^{*r}(\mathbf{x}^{(i)})z_i$, we know, for the optimal values of the two problems, $p^* \leq g^*$ holds from convex optimization. Equality $p^* = g^*$ holds if $p$, $g$ are convex, and the \emph{Slater constraint qualification} holds, which requires the problem to be strictly feasible. 
\item For any values of $B > 0$, if we consider very small values of some $z_i$, the strict inequality $\sum_{i=1}^n\tau^{*c}(\mathbf{x}^{(i)})z_i < B$ can always hold. Further, $B$ is usually large for a marketing campaign. Thus Slater qualifications hold. 
\end{itemize} 

From the analysis above, Problem~\ref{eq:pstatement_constrained} and its dual problem~\ref{eq:duality_problem} are equivalent, and we can solve Problem~\ref{eq:duality_problem} by iteratively optimizing with respect to $\mathbf{z}, \lambda$.  

\textbf{Optimize $\mathbf{z_i}$:} Keeping $\lambda, \mathbf{\tau}$ fixed, as $\lambda$ and $\ B$ are constants, Problem~\ref{eq:duality_problem} becomes: 
\begin{equation} 
\begin{split}
  \label{eq:lagrangian_reduced}
  \text{maximize}\quad \sum_{i=1}^nz_i s_i \\
  \text{subject to} \quad 0\leq z_i \leq 1
\end{split}
\end{equation} 
Where we define the \emph{effectiveness score} $s_i = \tau^{*r}(\mathbf{x}^{(i)})-\lambda\tau^{*c}(\mathbf{x}^{(i)})$. This optimization problem has a straightforward solution: assign the multiplier $z_i = 1$ when the ranking score $s_i \geq 0$ and assign $z_i = 0$ when ranking score $s_i < 0$. 

\textbf{Optimize $\lambda$:} Take the derivative of $L$ with regards to $\lambda$, $\frac{\partial g}{\partial \lambda}=B-\sum_{i=1}^n\tau^{*c}(\mathbf{x}^{(i)})z_i$. We can update $\lambda$ by Eq. (\ref{eq:update_lambda}) where $\alpha$ is the learning rate. 
\begin{equation}
  \label{eq:update_lambda}
  \lambda\rightarrow \lambda + \alpha(B-\sum_{i=1}^n\tau^{*c}(\mathbf{x}^{(i)}))
\end{equation}
Based on the two steps above, we can iteratively solve for both $z_i$ and $\lambda$ \cite{bertsekas1999nonlinear} .

In the next part, we solve for the $\tau^*$ functions, then finally connect components together to form the eventual algorithm. We can leverage quasi-oracle estimation of the CATE function $\tau$~\cite{nie2017quasi}. Concretely, the $m^*$ function, and optionally $e^*$ function, are fitted with L2 regularized linear regression, then $\tau^*$ functions are fitted with Eq.~\ref{eq:quasi_oracle_balance}. The problems are convex and have deterministic solutions. 

In our \emph{Duality R-learner} algorithm, we take an approach to combine the two $\tau^*$ functions into one model. Instead of learning $\tau^{*r}$ and $\tau^{*c}$ respectively, we fit a single \emph{scoring model} $s_i=\tau^{*E}(\mathbf{x}^{(i)})$ in Eq.~\ref{eq:lagrangian_score}. Note the Duality solution suggests we should include any sample with $\hat\tau^{*E}(x_i)>0$. Larger this value, more contribution the sample will have and thus a higher ranking it should get. 
\begin{equation} 
  \label{eq:lagrangian_score} 
  s_i=\tau^{*E}(\mathbf{x}^{(i)})=\tau^{*r}(\mathbf{x}^{(i)})-\lambda\tau^{*c}(\mathbf{x}^{(i)}) 
\end{equation} 
This form is linear, so we can use $Y^E=Y^r-\lambda Y^c$ instead of the the original $Y$ (single outcome for value and cost respectively) in the estimators above. Specifically, Eq. (\ref{eq:lagrangian_linearity}). 
\begin{align} 
  \label{eq:lagrangian_linearity} 
  \tau^{*r}(\mathbf{x}) - \lambda \tau^{*c}(\mathbf{x}) &= E((Y^r_1 - \lambda Y^c_1 - (Y^r_0  - \lambda Y^c_0)| \mathbf{X} = \mathbf{x}) \\&= E(Y^E_1 - Y^E_0 | \mathbf{X} = \mathbf{x}) 
\end{align} 
Then we train a regression model through the quasi-oracle estimation method, with this $Y^E$ and the output becomes $\tau^{*E}$ which could be used directly. This has two benefits: first, we optimize a joint model across $Y^r$ and $Y^c$ for the parameters to be able to find correlations jointly; second, for production and online service, we will arrive at one single model to perform prediction. 


We iteratively solve the Duality R-learner algorithm. This duality method lightens the production burden of having multiple models, and the algorithm can jointly improve cost and benefit by directly solving the constrained optimization problem for balanced effectiveness. 
\vspace{-0.2cm}
\subsection{Direct Ranking Model} 
The approach described in the previous section contains two separate steps, treatment effect prediction and constraint optimization. The ultimate business objective is to identify a \emph{portfolio of users} that we can achieve highest incremental user cross-sell or up-sell with a cost budget, which does not rely on the perfect individual prediction (point estimate) of treatment effect, but rather, achieves the overall market-wide effectiveness. This is similar to the search ranking algorithm to optimize for a holistic ranking objective vs Click Through Rate (CTR) point  estimate~\cite{huang2013learning}~\cite{shen2014a}. We aim to achieve better performance by combining these two steps together, and this is the algorithm we propose: Direct Ranking Model (DRM). 

This model tries to solve an unconstrained optimization problem where we minimize the cost per unit of gain: 
\begin{align} 
\label{eq:pstatement_unconstrained_opt} 
\text{minimize} \quad \frac{\bar{\tau}^{*c}(\mathbf{x})}{\bar{\tau}^{*r}(\mathbf{x})} 
\end{align} 


\textbf{Model and Effectiveness Objective.} We can then construct our model and the loss function as follow. In Eq. (\ref{eq:drm_1}) $f$ is the function the model will learn with tunable parameters. This function outputs an effectiveness score, indicating how efficient the sample is based on its features $\mathbf{x^{(i)}}$. $f$ can be in any differentiable form such as linear or a neural network structure. 
\begin{equation}
  \label{eq:drm_1} 
  S_i = f(\mathbf{x}^{(i)}) 
\end{equation} 
We use a standard hyperbolic tangent as non-linear activation for the neural network($\tanh$). 
\begin{equation} 
  \label{eq:drm_2}
  s_i = \tanh(S_i)
\end{equation}
We then normalize the effectiveness scores using the softmax function to arrive at $p_i$ for each user (Eq. (\ref{eq:drm_3_1})). $p_i$ sum to 1 in each cohort respectively, for $T_i=1$ and $T_i=0$. 
\begin{equation}
  \label{eq:drm_3_1}
  p_i = \frac{e^{s_i}}{\sum_{j=1}^n\mathbb{I}_{T_j=T_i}e^{s_j}}
\end{equation}
Here $\mathbb{I}_{T_j=T_i}$ is the indicator function for sample $j$ whether it's in the same group (treatment or control) as sample $i$. 
Based on this, we can calculate the expected treatment effect of our user portfolio. We can write effectiveness weighted sample treatment effect for retention and cost with (Eq. (\ref{eq:drm_4_1}), Eq. (\ref{eq:drm_4_2})).
\vspace{-0.1cm}
\begin{equation}
  \label{eq:drm_4_1}
  \bar\tau^{*r}=\sum_{i=1}^nY^{r(i)}p_i(\mathbb{I}_{T_i=1} - \mathbb{I}_{T_i=0})
\end{equation}
\vspace{-0.1cm}
\begin{equation}
  \label{eq:drm_4_2}
  \bar\tau^{*c}=\sum_{i=1}^nY^{c(i)}p_i(\mathbb{I}_{T_i=1} - \mathbb{I}_{T_i=0})
\end{equation}
\vspace{-0.1cm}
Finally, we have our loss function in Eq. (\ref{eq:drm_5}), which is the ratio of treatment effects as the holistic efficiency measure plus a regularization term.
\vspace{-0.1cm}
\begin{equation}
  \label{eq:drm_5}
  \hat{f}(\cdot ) = argmin_{f}\left \{ \frac{\bar{\tau^c}}{\bar{\tau^r}}+\Lambda_n(f(\cdot ))  \right \}
\end{equation}
\vspace{-0.1cm}
Since all the operations above are differentiable, we can use any off-the-shelf optimization method to minimize the loss function and learn the function $f$. Because the direct optimization is well suited for deep learning, we incorporated this method with the deep learning architectures and frameworks, and implemented our approach using TensorFlow \cite{abadi2016tensorflow} and used Adam optimizer \cite{kingma2014adam}. The definition of $f$ function is flexible for instance, multi-layer neural networks, convolutional and recurrent networks. 

\vspace{-0.2cm}
\subsection{Constrained Ranking Models} 

Constraints are inherent in retention and engagement products, such as a fixed cost budget or product limitations to send to only 30\% quantile of the users. Despite the previous model is able to directly optimize for market-wide effectiveness and utilize powerful deep learning models, the algorithm is disadvantaged with constraints and may not find the best solution. 

There is also difficulty in leveraging deep learning models to solve hard-constrained optimization problems~\cite{marquez2017imposing}. To address these difficulties, we develop methods to turn hard constraints into soft constraints applicable to the deep learning methodology. Concretely, we enable this by developing two novel deep learning components: \emph{Quantile pooling} and \emph{constraint annealing}.

\textbf{Quantile Pooling} Many deep learning algorithms apply the critical step of \emph{pooling}. Pooling applies a mathematical operator such as $max$ or $mean$ to selectively retain values from the previous layer. These operations create useful sparsity in deep learning architectures which eases pressure on the numerical optimization process and increase invariance in the top layer representations~\cite{lecun1995convolutional}~\cite{goodfellow2009measuring}~\cite{zou2012deep}~\cite{jarrett2009best}~\cite{le2011learning}~\cite{le2013building}. In this section, we describe the new pooling method for selecting a quantile of effectiveness measures from the whole population using a sorting operator. This pooling component enables us to systematically select output satisfying constraints and dynamically construct efficiency objective focused on those selections. We propose this method with the deep learning architecture in our causal learning framework.

We assume either a quantile $q\%$ or a cost budget $B$ is given as a fixed hyper-parameter. For the former, we are constrained to offer treatment to top $q\%$ of the users, for the latter, we could not exceed the budget $B$. 

Leveraging methodologies developed in the previous section (Eq.~\ref{eq:drm_1}, Eq.~\ref{eq:drm_2}), at optimization iteration $(k)$, for user $(i)$ in the dataset, users' effectiveness score is calculated as below. Assume $s_i^{(k)}$ is the original score: 
\vspace{-0.2cm}
\begin{align*} 
	s_i^{(k)} = \tanh (f^{(k)}(\mathbf{x}^{(i)})) 
\end{align*} 
\vspace{-0.1cm}
The treatment decision depends on the value of $s_i^{(k)}$ and its mathematical relationship with our constraints. We abstract this treatment decision with a fall-off function $\sigma^*$ (chosen to be a \emph{sigmoid function}) and an input offset $d^{*(k)}$, shown in Eq.~\ref{eq:offset_fall_off}. It illustrates how this offset $d$ lead to a fall-off variable $v$ which discounts output scores. In this equation the $t^*$ variable is a hyperparameter called \emph{temperature} to control softness of the fall-off.
\vspace{-0.2cm}
\begin{align} 
\label{eq:offset_fall_off} 
	V_i^{(k)} = s_i^{(k)} - &d^{*(k)} \quad v_i^{(k)} = \sigma^* (V_i^{(k)}) = \frac{1}{1 + \exp(-t^*V_i^{(k)})}
\end{align} 
\vspace{-0.1cm} 
Here the offset $d_i$ is determined by both constraints and the population of scores $\mathbf{s}^{(k)}=\{s_i^{(k)}\}$ at iteration $(k)$. In this paper, we give two definitions of this offset transform: 

\textbf{\emph{Top Quantile Constraint}}: For optimization constrained to a fixed quantile $q\%$, we related the offset with a quantile function $\Gamma_Q(\mathbf{s^{(k)}}, q)$ where $q\%$ is the quantile percentage above which we decide to offer treatment:  
\vspace{-0.1cm}
\begin{align*} 
    d^{*(k)} = \Gamma_Q (\mathbf{s}^{(k)}, q) 
\end{align*} 
\vspace{-0.1cm}
The $\Gamma_Q$ function is implemented using a \emph{sorting operator} $\Psi$ and \emph{take $n$th} operator $\mu$, and $n = N\frac{q}{100}$ where N is total number of users in the population: 
\vspace{-0.1cm}
\begin{align*} 
	\Gamma_Q (\mathbf{s}^{(k)}, q) = \mu(\Psi(\mathbf{s}^{(k)}), n) 
\end{align*} 
\vspace{-0.1cm}
Semantically it means we first sort user effectiveness scores then take the q\% quantile value as offset $d^{(k)}$. 

\textbf{\emph{Fixed Cost Constraint}}: For optimization constrained to a fixed cost $B$, we related the offset with a cost-limiting function $\Gamma_C(\mathbf{s^{(k)}}, B)$:  
\vspace{-0.1cm}
\begin{align*} 
    d^{*(k)} = \Gamma_C (\mathbf{s}^{(k)}, B) 
\end{align*} 
\vspace{-0.1cm}
Similarly, the $\Gamma_C$ function is implemented using the \emph{sorting operator} $\Psi$ and \emph{cumulative sum} operator $\omega$, and a operator $\Phi$ that represents a function which returns the effectiveness score corresponding to the input's last element that's smaller than $B$: 
\vspace{-0.1cm}
\begin{align*} 
	\Gamma_Q (\mathbf{s}^{(k)}, B) = \Phi(\omega(\Psi(\mathbf{s}^{(k)})), B, \mathbf{s}^{(k)})
\end{align*} 
\vspace{-0.1cm}
Semantically, we sort users based on their effectiveness scores, then take quantile value of $\mathbf{s}^{(k)}$ as offset $d^{(k)}$, where the quantile value corresponds to the rank of user just before where the budget exceeds $B$. 

Despite the sophistication of these definitions, all the operators defined are differentiable, thus can be easily incorporated into the deep learning framework. This \emph{Quantile Pooling} mechanisms \emph{deactivates} or \emph{nullify} outputs that do not satisfy constraints with the equation below: 
\vspace{-0.2cm}
\begin{align} 
\hat{s}_i^{(k)} = s_i^{(k)} \sigma^*(s_i^{(k)} - d^{*(k)})
\end{align} 
\vspace{-0.1cm}
The intuition for quantile pooling is analogous to max-pooling. The model dynamically creates sparse connection patterns in the neural network to focuses on the largest activations across a population of neurons. This algorithm structures the model for reducing against model variance and helps optimizers to find better local minima. 

We replace the effectiveness score $s_i$ in Eq.~\ref{eq:drm_3_1} with the score after pooling $\hat{s}_i$. The quantile pooling ensures on every optimization iteration, the eventual effectiveness objective is focused on users that are valid according to constraints. Finally the constraints are soft, so we translate constraints into the architecture of the model, and the user effectiveness scoring function is eventually learned through direct and unconstrained optimization. 

\textbf{Constraint annealing} The temperature term $t^*$ in Eq.~\ref{eq:offset_fall_off} determine how hard the fall-off function is, thus determines the hardness of constraints. We observed difficulties optimizing the model with constrained ranking when $t^*$ is set large and constraint is hard. The early stages of optimization could not find local minima because the gradients are small with a sharp cut-off sigmoid. At the same time, when we set $t^*$ too small, the performance is similar to Direct Ranking Model (Eq.~\ref{eq:drm_5}). 

We propose an annealing process on the parameter $t^*$ to have a schedule of rising temperature~\footnote{The exact annealing parameters are in the Empirical Results section.}. This allows gradient methods for optimization to be effective at early stages of optimization, and when the model settles in a better local minima, the constraints could be tightened so solutions that fit into those constraints could be found. 


\subsection{Evaluation Methodology} 
The business objective is to achieve most incremental user retention with a given  cost budget. The retention and cost here are two critical values to trade-off. 

\textbf{\emph{Cost Curve}}. With two treatment outcome $\tau^{r}$ and $\tau^{c}$, we draw a curve and use cost as X-axis and retention as Y-axis as the illustration below. 
\vspace{-0.1cm}
\begin{figure}[h] 
  \centering 
  \includegraphics[width=\linewidth]{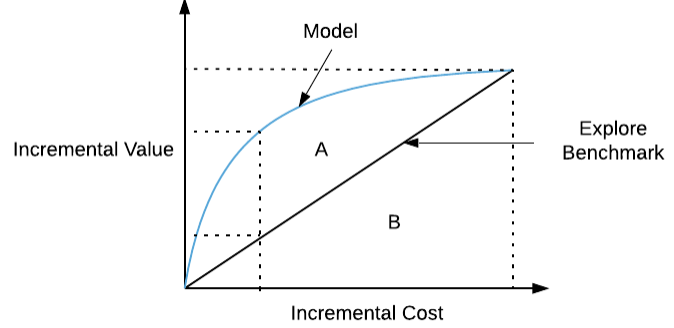} 
  \caption{Illustration of the Cost-Curve.} 
  \label{cost_curve_illustration} 
\end{figure} 
Samples are ordered by the effectiveness score $S_i = f(X_i)$ on the cost curve. For each point on the curve, we take the number of treatment samples at this point on the curve, multiplied by ATE (Average Treatment Effect) of this group. 
\begin{align*}
\#\{T_i = 1 | S_i > S_i^{pth}\} \times ATE(x_i | S_i > S_i^{pth})
\end{align*}
Therefore each point represents aggregated incremental cost and value, usually both increasing from left to right. From origin to right-most of the curve, points on the curve represents the outcome if we include $p\%$ of the population for treatment, $p\in[0, 100]$.


If the score $S_i$ is randomly generated, the cost curve should be a straight line. If the score is generated by a good model, then the curve should be above the benchmark line, meaning for the same level of incremental cost, samples are selected to achieve higher incremental value. 

\textbf{\emph{Area Under Cost Curve (AUCC)}}. Similar to Area Under Curve of ROC curve, we define the normalized area under cost curve as  the area under curve divided by the area of rectangle extended by maximum incremental value and cost, or the area ratio $\frac{A + B}{2B}$. A and B are the area shown in the cost curve figure. So the AUCC value should be bounded within [0.5, 1) and larger the AUCC, generally better the model. 

\section{Empirical Results} 
\label{sec:empirical_results} 

In this section, we will cover the empirical results to compare proposed algorithms with prior art approaches (Causal Forest, R-Learner) on marketing and pubic datasets. We will first describe the experiment data set and experiment setup. Then we would analyze both offline and online test results. In summary, cost curve offline evaluation is consistent with real-world online result and our proposed methods perform significantly better versus previous methods. 

\subsection{Experiments} 
The application goal of our model is to rank users from most effective to least effective, so that the overall market-wide metrics are optimized. As stated in algorithm section, we train our model on data logged from previous experiments with treatment assignment logs and the actual outcomes. 

\subsubsection{Experiment with Marketing Data} 
\label{sec:explore_exploit_exp} 
We adopt an explore and exploit experimental set-up in the paradigm of reinforcement learning~\cite{liu2018explore} and multi-armed bandits~\cite{katehakis1987multi}~\cite{li2010contextual}. We launch experiment algorithms in a cyclic fashion. For each cycle we have 2 experiments: explore and exploit, which contains non-overlapping sets of users. The explore experiment is randomly given, and serves to collect data from all possible scenarios. On the other hand, exploit applies model and other product features to optimizes performance. The experiment design is illustrated in the following chart.  
\begin{figure}[h] 
  \centering 
  \includegraphics[width=\linewidth]{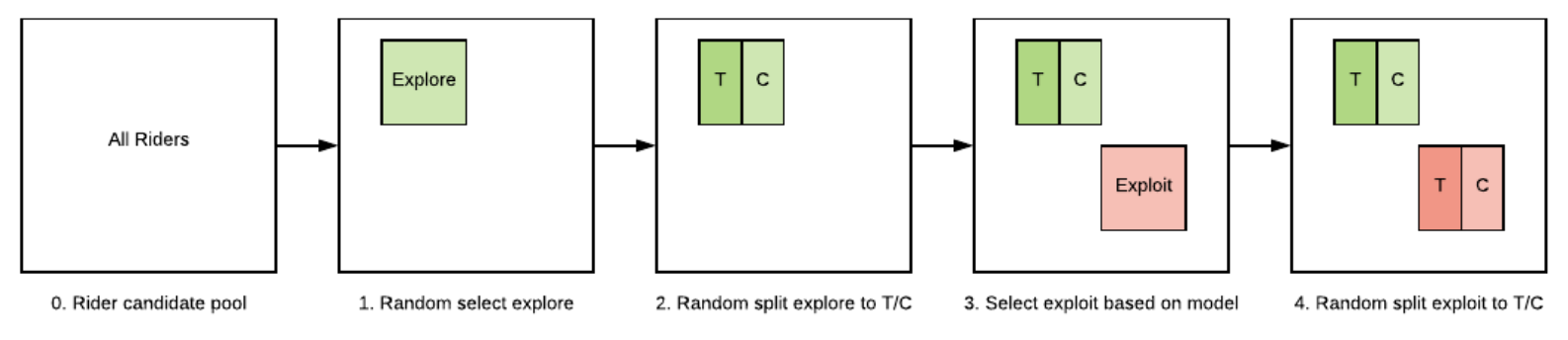} 
  \label{explore_exploit_chart} 
\end{figure} 

\emph{Explore}. Users are randomly selected without any product specific algorithm, into explore experiments from the predefined user candidate pool. This allows us to collect an unbiased dataset which represents the whole population. Once users are selected, we then randomly give treatment / control assignment with a fixed probability\footnote{The number of samples in explore is solely determined by the budget.}. 

\emph{Exploit}. Excluding users already in explore experiments, based on model and budget we select users into exploit experiments. This exploit group is for product application. 



We use explore for model training and offline performance evaluation and exploit for online performance evaluation. 
We collect data from experiments following this design. For each sample, we will log their feature constructed with data before experiment starts, experiment label (explore or exploit), treatment control assignment and outcomes (value and cost). Outcomes are aggregated within the experiment period. Value outcome could be any arbitrary desired business value the specific definition of which is unrelated to the algorithm, while cost outcome is also arbitrary undesired cost. 

\textbf{Marketing Data} To obtain data for model training and offline evaluation, we utilize a randomized \emph{explore} online experiment. We first randomly allocating users to control and treatment cohorts (A/B). For the treatment cohort, we give all users treatment.  In this experiment we collected millions of user level samples in multiple experiment periods. Following is an illustrative table for the dataset we collected. 

\begin{table} 
  \caption{Example marketing dataset}
  \label{tab:example_data}
  \begin{tabular}{llllll}
    \toprule
    user id & strategy & $X_i$ & $T_i$& $Y^r$ & $Y^c$ \\
    \midrule
    A & explore & $(1.2, 3, ...)$ & $1$& $3$ & $2.3$ \\
    \midrule 
    B & exploit & $(2.4, 1, ...)$ & $0$& $1$ & $0.1$ \\ 
    \bottomrule
\end{tabular}
\end{table}



\vspace{-0.3cm}
\subsubsection{Causal Experiments Designed with Public Datasets} 

The effectiveness of our proposed causal models is mainly experimented with marketing data. To ensure reproducibility we also experiment on public datasets. We make assumptions to select treatment assignment and outcomes on available data vectors to design simulated experiments for our proposed causal models. 

\textbf{US Census 1990} The US Census (1990) Dataset (Asuncion \& Newman, 2007~\cite{uscensulink} contains data for people in the census. Each sample contains a number of personal features (native language, education...). The features are pre-screened for confounding variables, we left out dimensions such as other types of income, marital status, age and ancestry. This reduces features to d = 46 dimensions. Before constructing experiment data, we first filter with several constraints. We select people with one or more children (\emph{`iFertil'} $\leq$ 2), born in the U.S. (\emph{`iCitizen'} = 0) and less than 50 years old (\emph{`dAge'} $<$ 5), resulting in a dataset with $225814$ samples. We select `treatment' label as whether the person works more hours than the median of everyone else, and select the income (\emph{`dIncome1'}) as the gain dimension of outcome for $\tau_r$, then the number of children (`iFertil') multiplied by $-1.0$ as the cost dimension for estimating $\tau_c$. The hypothetical meaning of this experiment is to measure the cost effectiveness, and evaluate who in the dataset is effective to work more hours.

\textbf{Covertype Data} The Covertype Dataset (Asuncion \& Newman, 2007) contains the cover type of northern Colorado forest areas with tree classes,  distance to hydrology, distance to wild fire ignition points, elevation, slope, aspect, and soil type. We pre-filter and only consider two types of forests: \emph{`Spruce-Fir'} and \emph{`Lodgepole Pine '}, and use data for all forests above the median elevation. This results in a total of $247424$ samples. After processing and screening for confounding variables, we use 51 features for model input. With the filtered data, we build experiment data by assuming we are able to re-direct and create water source in certain forests to fight wild fires, but also like to ensure the covertype trees are not imbalanced by changing the hydrology with preference to `Spruce-Fir'. Thus, the treatment label is selected as whether the forest is close to hydrology, concretely, distance to hydrology is below median of the filtered data. The gain outcome is a binary variable for whether distance to wild fire points is smaller than median, and cost outcome is the indicator for `Lodgepole Pine' (1.0, undesired) as opposed to `Spruce-Fir' (0.0, desired). 

Marketing data, public US Census and Covtype datasets are split into 3 parts: train, validation and test sets with respective percentages 60\%, 20\%, 20\%. We use train and validation sets to perform hyper-parameter selection for each model type. The model is then evaluated on the test set. 

\subsubsection{Model implementation details} In this section we briefly give the implementation details of our models. 

\emph{\textbf{Quasi-oracle estimation (R-Learner)}}. We use Linear Regression\footnote{Using SKLearn library's Ridge Regression with 0.0 as the regularization weight.} as the base estimator. Since the experiment cohorts are randomly selected, we use constant treatment percentage as propensity in the algorithm. Since we need to define one CATE function for R-learner, we use the R-learner to model the gain value incrementality with $\tau_r$. 

\emph{\textbf{Causal Forest}}. We leverage the generalized random forest (\emph{grf}) library in R~\cite{wager2018estimation}~\cite{grflink}~\cite{athey2016recursive}. For details, we apply causal forest with 100 trees, 0.2 as alpha, 3 as the minimum node size, and 0.5 as the sample fraction. We apply the ratio of two average treatment effect functions in ranking by training two causal forests. To rank users or other cardinalities with respect to cost vs gain effectiveness, we estimate the conditional treatment effect function both for gain ($\tau_r$) and cost ($\tau_c$), i.e. train two Causal Forest models. For evaluation, we compute the ranking score according to the ratio of the two $\frac{\tau_r}{\tau_c}$. 
For hyper-parameters, we perform search on deciles for parameters \emph{num\_trees}, \emph{min.node.size}, and at 0.05 intervals for \emph{alpha}, \emph{sample.fraction} parameters. We also leverage the \emph{tune.parameters} option for the \emph{grf} package, eventually, we found best parameters through best performance on validation set\footnote{Best parameters are the same for all three datasets we experimented: \emph{num\_trees}$=100$ (50 trees for each of the two CATE function, $\tau_r$, $\tau_c$), \emph{alpha}$=0.2$, \emph{min.node.size}$=3$, \emph{sample.fraction}$=0.5$}. 

\emph{\textbf{Duality R-learner}}. Similar to R-learner, we use Ridge Regression as the base estimator and constant propensity, and apply the model stated in Eq.~\ref{eq:update_lambda} for ease of online deployment. The iterative process to solve $\lambda$ in Eq.~\ref{eq:lagrangian_score} is inefficient as the value function $g$ here is piece-wise linear w.r.t $\lambda$. Since Ridge Regression is lightweight to train, in practice, we take the approach to select $\lambda$ with best performance on the validation set.\footnote{We determine the value of $\lambda$ through hyper-parameter search on deciles and mid-deciles, e.g. $\lambda\in \{0.001, 0.005, 0.01, 0.05\}$; best $\lambda$ for marketing data is $0.1$, for US Census and Covertype data is $0.05$.} 

\emph{\textbf{Direct Ranking}}. We implement our deep learning based models with Tensorflow~\cite{abadi2016tensorflow}. To align with baseline and other methods in our experiments, we use a one layer with linear parameterization $\tanh(\mathbf{w}^T \mathbf{x} + b)$ as the scoring function, without weight regularization, the objective functions stated in the algorithm section are used. We use the Adam optimizer with learning rate 0.01 and default beta values. We compute gradients for entire batch of data, and run for 600 iterations. 

\emph{\textbf{Constrained Ranking}}. We experiment with the \emph{Top Quantile}  operator. In addition to using a linear scoring function, we use a consistent quantile target at 40\%, and apply a starting sigmoid temperature of $0.5$, and use constraint annealing at increments of 0.1 every 100 steps of Adam optimizer. For constraint annealing, we validate and select different schedules. We note the quantile pooling offers a flexible lever $d^{(k)}$ to minimize objective function, making the optimization unstable. We stop the gradient on $d^{(k)}$ to disable the fast changing of this value. 

\vspace{-0.2cm}
\subsection{Results on Causal Learning Models} 
\begin{figure}[h] 
  \includegraphics[width=0.85\linewidth]{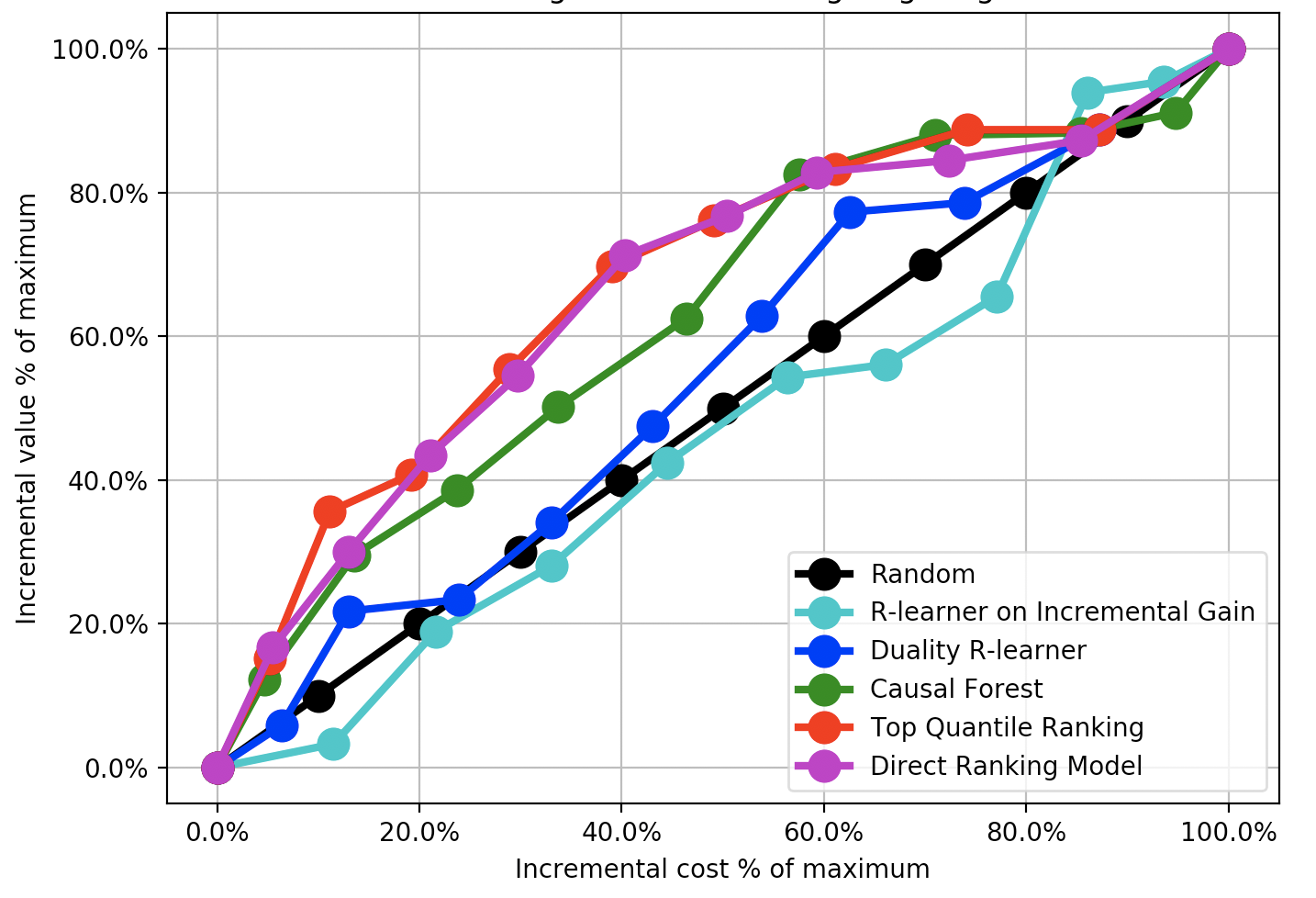} 
  \caption{Cost-Curve results for marketing data.} 
  \label{marketing_data_result} 
\end{figure}

Figure~\ref{marketing_data_result} shows the cost curve  for each model on marketing data test set. The baseline R-learner optimized for incremental gain $\tau_r$ could not account for the cost outcome and under-performs on our task. Thus we use Duality R-learner as a benchmark for all our experiments. Causal Forest also perform reasonably well. Direct Ranking out-performs previous models with \emph{22.1\%} AUCC improvement upon Duality R-learner, and Constrained Ranking algorithm is the best performing model on the marketing dataset, out-performing Duality R-learner by \emph{24.6\%} in terms of AUCC.

\begin{figure}[h] 
  \includegraphics[width=0.75\linewidth]{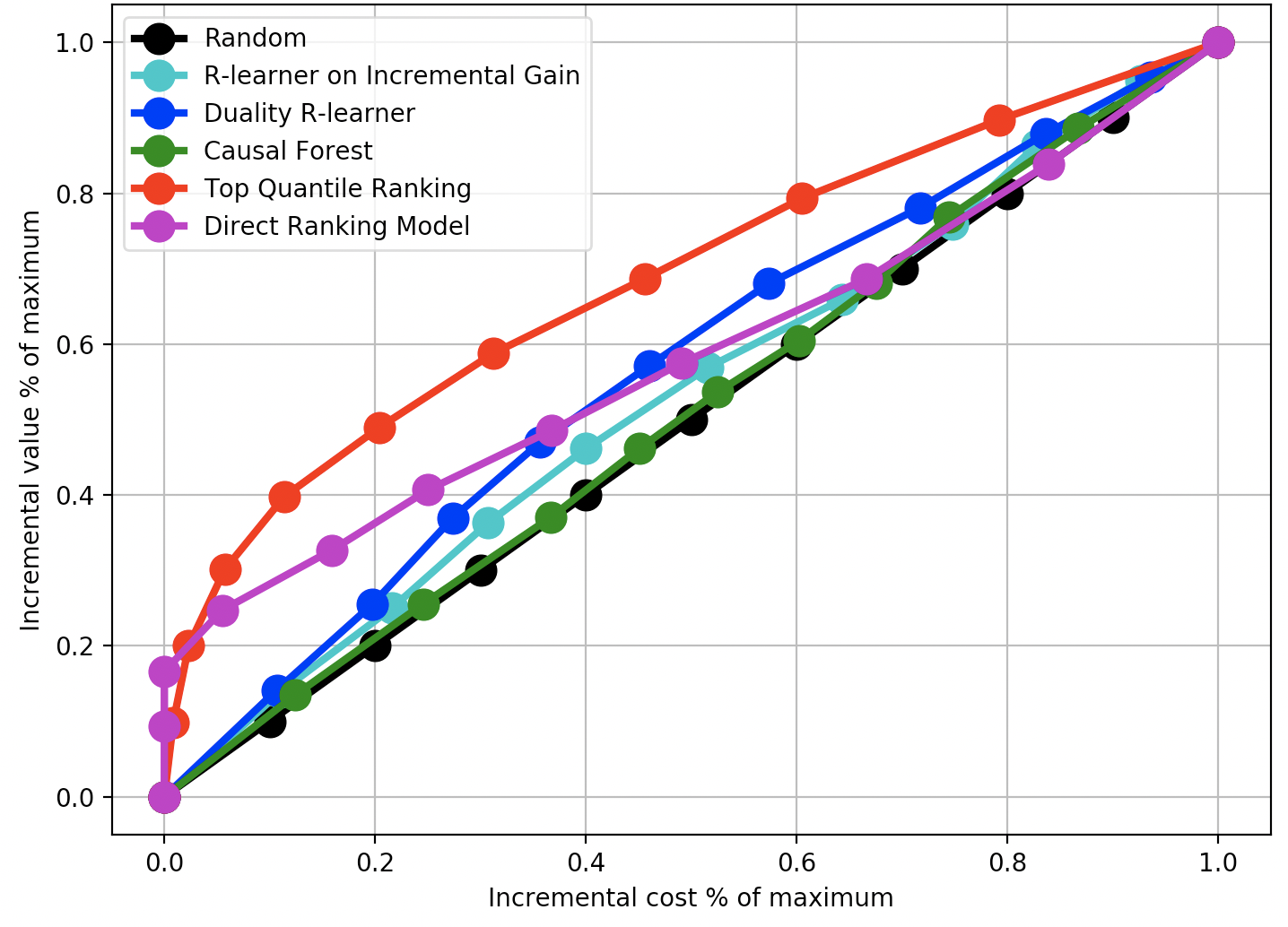} 
  \caption{Cost-Curve results for public US Census data.} 
  \label{fig:uscensus_result} 
\end{figure}

\begin{figure}[h] 
  \includegraphics[width=0.75\linewidth]{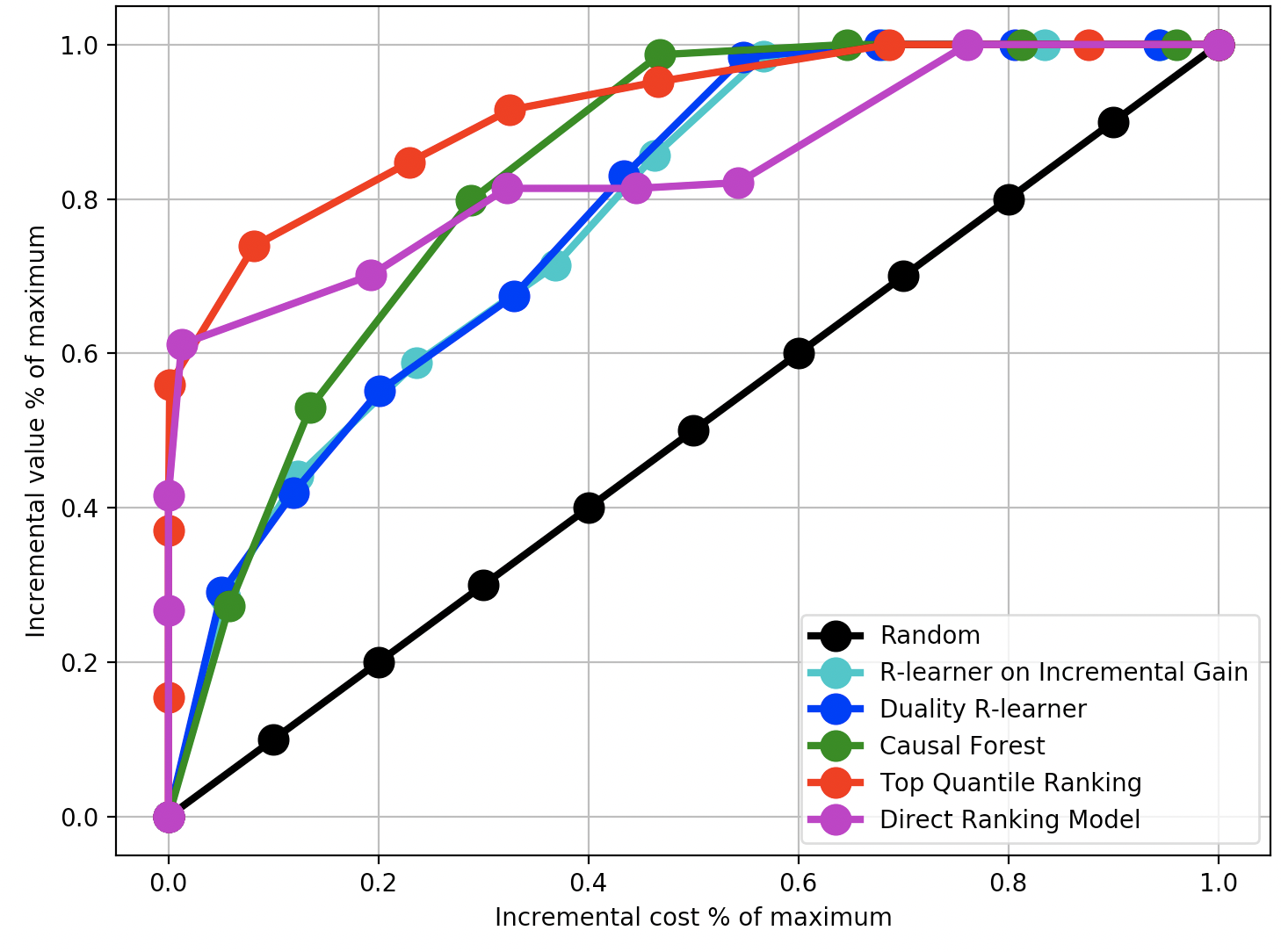} 
  \caption{Cost-Curve results for public Covtype data.} 
  \label{fig:covtype_result}
\end{figure}

Figure~\ref{fig:uscensus_result} shows results of causal models on US Census. The baseline R-learner on gain performs slightly better due to less cost impact. Duality R-learner still works reasonably well. Direct Ranking and Constrained Ranking out-performs Duality R-learner by \emph{2.8\%} and \emph{21.2\%}, respectively to AUCC \emph{0.58} and \emph{0.69}. 
Figure~\ref{fig:covtype_result} shows results of causal models on Covtype datasets. The optimization problem on this dataset is easier as results on multiple models are better. Direct Ranking and Constrained Ranking out-performs Duality R-learner by \emph{7.3\%} and \emph{17.9\%}, with AUCC for Constrained Ranking algorithm as high as \emph{0.92}. 

Table~\ref{tab:summary_result_table} shows results Constrained Ranking and Direct Ranking algorithms are significantly better, more than \emph{25\%} in terms of AUCC, than R-learner on gain outcome, and out-performs Duality R-Learner by around \emph{10\%}. One example for cost effectiveness is to look at the vertical dash line at $20\%$ of total incremental cost, we can achieve 2X more incremental retention than random selection by using our causal models. This can result in 50\% reduction in cost. 

\begin{table}
  \caption{Summary of AUCC results across models and datasets.} 
  \label{tab:summary_result_table}
  \begin{tabular}{llllll}
    \toprule
    Algorithm & Prod. &\% imp. & USCensus &Covtype \\ 
    \midrule 
    Random & 0.500 && 0.500 & 0.500 \\
    \midrule 
    R-learner G & 0.464 && 0.533 & 0.779 \\
    Duality R-learner & 0.544 &0.0\%& 0.567 & 0.783 \\
    Causal Forest & 0.628 &15.4\%& 0.510 & 0.832 \\
    Direct Ranking & 0.664 &22.1\%& 0.583 & 0.840 \\
    Constrained Ranking & \textbf{0.678} &24.6\%& \textbf{0.687} & \textbf{0.915} \\
    \bottomrule
\end{tabular}
\end{table} 

The models we proposed has been deployed in production and operates in multiple regions. The model is developed using data from previous experiments, and launched as a prediction model in production to rank users. In this section we describe the challenges and learning of putting the model in production. 

\begin{figure}[h] 
  \centering 
  \includegraphics[width=\linewidth]{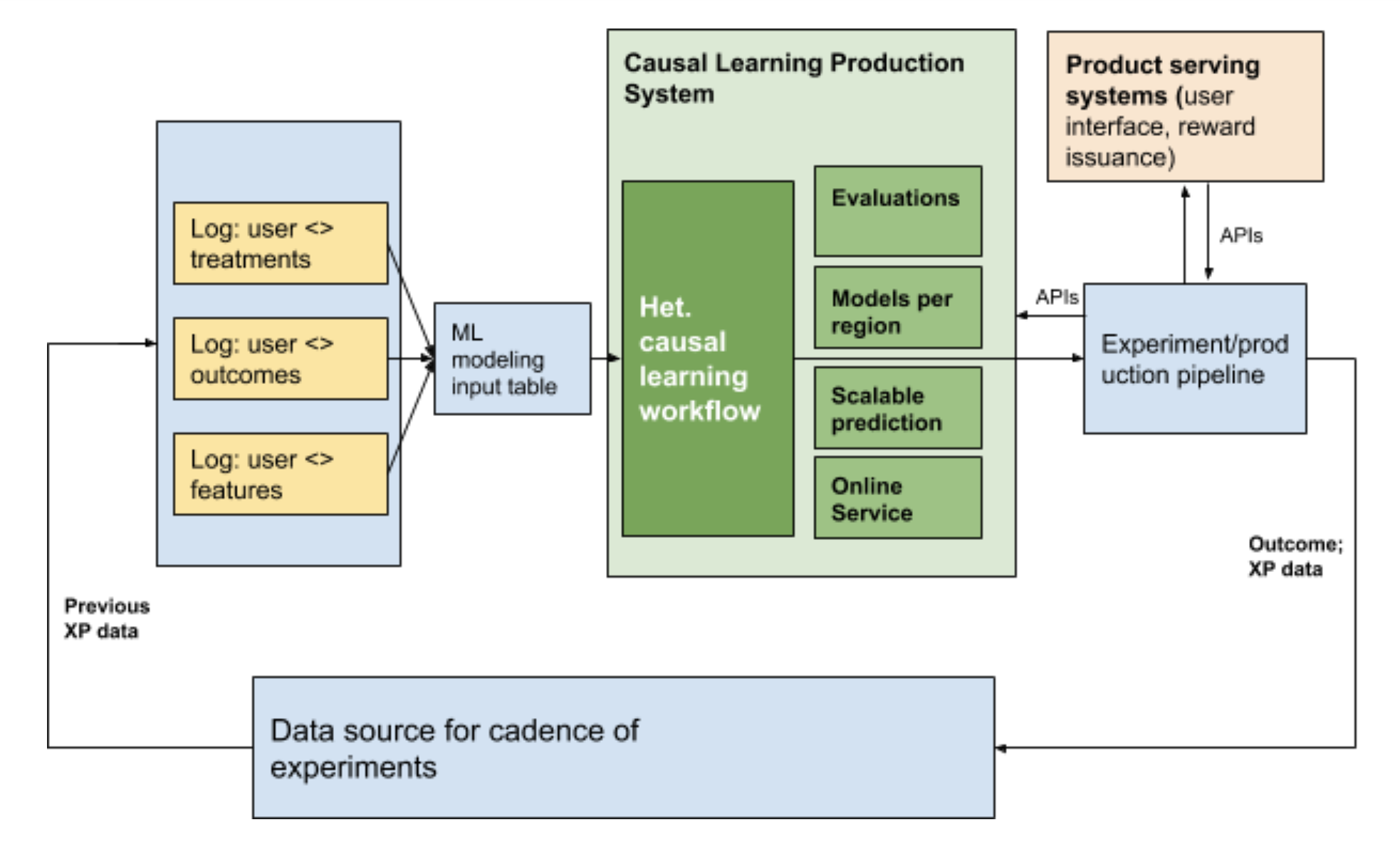} 
  \caption{Production system for causal learning.} 
  \label{fig:kdd_prod_eng} 
\end{figure}

\emph{\textbf{Engineering system for production}} The first challenge is designing an engineering system to support our causal learning approach. Different from traditional machine learning models, we build a Heterogeneous Causal Learning Workflow (HCLW) system to learn from observed outcomes of predefined treatments in previous launches. The previous production launch offers training data for subsequent launch, thus the product will improve decisions and model settings across a sequence of launches before forming the eventual production model. The design of this engineering system is shown in Figure~\ref{fig:kdd_prod_eng}. The data are collected from previous launches in the form of Table~\ref{tab:example_data}, and stored in offline storage before feeding into the causal learning pipeline. The pipeline produces the trained model, evaluation, and service components. Service components offers model decisions on users, interact with launch pipelines and product serving system through API's to issue rewards through user interface in the production system. 

\emph{\textbf{Production and offline evaluation}} Another important consideration to deploy to production is alignment of evaluation results. Unlike the full cost-curve metrics for offline evaluation, in online case, we could only measure one specific point on the cost curve for each model. The slope of the straight line between that point and origin measures the general cost effectiveness. This slope is given as $R$ in Eq.\ref{eq:slope_r}. If both models have similar spend level (similar value on x-axis), this slope would sufficiently capture the performance. 

\vspace{-0.1cm}
\begin{equation}
  \label{eq:slope_r}
  R = \frac{ATE^r(x_i|selected)}{ATE^c(x_i|selected)}
\end{equation}
\vspace{-0.1cm}

As mentioned in Section~\ref{sec:explore_exploit_exp}, we have both explore and exploit in our online setup. In this case when we try to test 2 models for comparison between DRM and Causal Forest, we will have 1 explore (random selection) and 2 exploits (model based selection). Within selected users we then random split them into treatment and control. To make the numerical metric uniform, we use $R$ for explore as benchmark and derive Eq.~\ref{eq:ratio_r}, which represents the relative efficiency gain compared to the benchmark. 

\vspace{-0.1cm}
\begin{equation}
  \label{eq:ratio_r}
  \frac{R_{exploit}-R_{explore}}{R_{explore}}
\end{equation}
\vspace{-0.1cm}


Online results are consistent with the offline results that all models perform  significantly better than explore and DRM consistently out-performs quasi-oracle estimation (R-learner) and Causal Forest.

\section{Conclusion and Future Work} 

\subsection{Conclusion} 
We propose a novel ranking method to optimize heterogeneous treatment effect for user retention. The method combines prediction and optimization into one single stage and provides a loss layer that can be incorporated with any deep learning structure. We also provide an empirical evaluation metric and adjustments for existing estimator for the  treatment effect optimization problem. We evaluate various methods empirically both offline and online. Our proposed method achieves significantly better performance than explore benchmark and existing estimators. After successful test, this method has been deployed to production and is live in many regions all over the world.

\subsection{Future Work} 
\emph{Smart Explore/Exploit}. In current work we use epsilon-greedy explore, where we split a fixed percentage of budget to spend on fully randomized explore to collect data for model training. However, this will sacrifice the overall performance and is suboptimal. As a better approach, we will try to use multi-arm bandit or Bayesian optimization framework to guide our smart explore based on the model uncertainty. 

\emph{Deep Embedding}. Raw time and geo features are extremely sparse. Various embedding techniques have been used for sparse features but none of them is specifically for treatment effect. As treatment effect is different from its underlying outcome, the embedding should also be different. Now that we have a general loss layer which could be incorporated with any deep learning structure, we could start to work on the embeddings specifically for treatment effects. 


\bibliographystyle{ACM-Reference-Format}
\bibliography{main_kdd}

\end{document}